%% file: main.tex
\documentclass[10pt,twoside,journal]{IEEEtran} 

\usepackage{amsfonts}
\usepackage{bm}
\usepackage{booktabs}
\usepackage{colortbl} 
\usepackage{xcolor}
\usepackage{subfigure}
\usepackage{multirow}
\usepackage{amssymb}
\usepackage{amsmath}
\usepackage{balance}

\ifCLASSOPTIONcompsoc
  \usepackage[nocompress]{cite}
\else
  \usepackage{cite}
\fi

\ifCLASSINFOpdf
   \usepackage[pdftex]{graphicx}
   \graphicspath{{../pdf/}{../jpeg/}}
   \DeclareGraphicsExtensions{.pdf,.jpeg,.png}
\else
   \usepackage[dvips]{graphicx}
   \graphicspath{{../eps/}}
   \DeclareGraphicsExtensions{.eps}
\fi

\usepackage{algorithm}  
\usepackage{algorithmic}
\usepackage{array}
\ifCLASSOPTIONcompsoc
 \usepackage[caption=false,font=footnotesize,labelfont=sf,textfont=sf]{subfig}
\else
 \usepackage[caption=false,font=footnotesize]{subfig}
\fi
\ifCLASSOPTIONcaptionsoff
 \usepackage[nomarkers]{endfloat}
\let\MYoriglatexcaption\caption
\renewcommand{\caption}[2][\relax]{\MYoriglatexcaption[#2]{#2}}
\fi
\let\MYorigsubfloat\subfloat
\renewcommand{\subfloat}[2][\relax]{\MYorigsubfloat[]{#2}}
\usepackage{url}

\begin{document}

\title{Beyond Sunny Skies: Using Large Language Models to Predict Urban Traffic in Adverse Weather}

\author{
    Zhu~Xiao,~\IEEEmembership{Senior Member,~IEEE},
    Kuan~Feng,
    Tong~Li,~\IEEEmembership{Member,~IEEE}, \\
    Chenxi~Liu, Gang~Wang, Zhuo~Tang, and Kenli~Li
    
    \IEEEcompsocitemizethanks{\IEEEcompsocthanksitem 
    Manuscript received XX XX 2026.
    
    Z.~Xiao, K.~Feng, T. Li, Z.~Tang, and K.~Li are with the College of Computer Science and Electronic Engineering, the Ministry of Education Key Laboratory of ``Fusion Computing of Supercomputing and Artificial Intelligence'', Hunan University, Changsha 410082, China.  E-mail:~\{zhxiao; thekuan; tliay; ztang; lkl\}@hnu.edu.cn.
    \IEEEcompsocthanksitem C.~Liu is with Hong Kong Institute of Science \& Innovation, Chinese Academy of Sciences, Hong Kong.~Email:~chenxi.liu@cair-cas.org.hk.
    \IEEEcompsocthanksitem G.~Wang is with CETC Key Laboratory of Aerospace Information Applications, Shijiazhuang, China.~Email:~wanggg@tju.edu.cn.}  
}

\IEEEtitleabstractindextext{%
\begin{abstract}
Urban traffic flow prediction is a cornerstone of intelligent transportation systems, yet accurate forecasting under adverse weather conditions remains a significant challenge. Existing data-driven models, while effective for routine patterns, often struggle with the sparse data and complex non-linear shifts associated with rare meteorological events. These methods typically treat environmental contexts as shallow numerical features, failing to capture the underlying semantic reasoning—why and how specific adverse conditions alter mobility behaviors across regions with different functional compositions. To address this limitation, we propose Multi-STG, a unified framework that integrates Large Language Model (LLM)-derived semantic priors with graph-based spatiotemporal learning. Multi-STG introduces an LLM-empowered semantic enhancement module that transforms comprehensive context descriptions such as weather, POIs, and holidays into high-level semantic embeddings, effectively providing a knowledge-driven complement to data scarcity. Furthermore, we design a multimodal attentional fusion mechanism to dynamically weigh heterogeneous inputs and a semantic-enhanced prediction head with gated residual calibration to robustly adjust forecasts based on semantic guidance. Extensive experiments on real-world multi-city private car datasets demonstrate that Multi-STG outperforms state-of-the-art baselines, reducing MAE by 11.82\% on average relative to the strongest baseline across three cities, while maintaining stable performance under sparse and non-stationary conditions.
\end{abstract}

}

\maketitle
\IEEEdisplaynontitleabstractindextext

\IEEEpeerreviewmaketitle

\input{Sec1_introduction}
\input{Sec2_relatedwork}
\input{Sec3_preliminaries}
\input{Sec4_method}

\input{Sec5_evaluation}

\section{Conclusion}
\label{sec:con}
In this paper, we present Multi-STG, a unified framework that integrates LLM-derived semantic priors with spatiotemporal graph learning to improve regional private vehicle flow prediction under adverse weather conditions. The framework combines regional feature modeling, LLM-based semantic enhancement, multimodal attentional fusion, a Graph Transformer backbone, and a semantic prediction head with gated residual calibration. Experiments on private vehicle datasets from three cities demonstrate improved forecasting accuracy, while sensitivity analyses indicate relatively stable performance across the tested hyperparameter settings. Dependence on LLM availability, potential semantic bias, and additional computational costs remain practical considerations. Future work can explore lightweight adaptation, fusion that accounts for semantic uncertainty, and model distillation for efficient deployment. These findings support the potential of combining contextual knowledge from language models with spatiotemporal learning for urban traffic forecasting.

\bibliographystyle{IEEEtran}

\bibliography{bibliography}

\end{document}

%% file: Sec1_introduction.tex
\section{Introduction}
Urban traffic flow prediction estimates future traffic conditions, such as volume, speed, and density, across a citywide road network by modeling historical mobility patterns and spatial dependencies~\cite{li2022dmgan, rong2021inferring, liu2026multi, liu2025vehicle}. As a core component of intelligent transportation systems, traffic flow prediction supports congestion management, dynamic route planning, and sustainable infrastructure development~\cite{diao2021impacts, hu2023spatio}. Accurate forecasts enable urban planners and traffic authorities to optimize signal control, allocate road resources more efficiently, and provide travelers with reliable guidance~\cite{wang2026joint}. However, urban traffic dynamics are strongly influenced by external factors, including adverse weather conditions (e.g., heavy precipitation and fog) and special events (e.g., holidays), which can cause abrupt, nonlinear departures from regular traffic patterns. Accurate, robust forecasting therefore requires capturing intrinsic spatiotemporal dependencies and integrating heterogeneous contextual information to account for traffic variations induced by these external factors~\cite{wang2020resilience, li2024causal}.

Extensive research has examined spatiotemporal urban traffic forecasting from various perspectives. Models such as LSTM~\cite{hochreiter1997lstm} and Transformer~\cite{vaswani2017attention} capture temporal dependencies, while graph neural networks~\cite{jiang2022gnnsurvey, zhang2021multi} model spatial relationships among road segments or regions. To account for external influences, multimodal approaches incorporate auxiliary information into forecasting. For example, T-GCN and MegaCRN capture non-periodic traffic shifts through temporal graph convolution and meta-graph learning, respectively~\cite{zhao2019tgcn, jiang2023megacrn}. Other methods incorporate meteorological observations, such as rainfall and visibility~\cite{koesdwiady2016improving, zhong2024weather}, or use points of interest to characterize regional functions~\cite{wang2023urban, li2023learning, xi2022beyond}. Despite these advances, integrating external information does not necessarily capture the mechanisms through which environmental conditions reshape traffic patterns. Existing approaches often incorporate auxiliary signals through direct encoding or attention mechanisms, but the resulting numerical representations may not adequately capture interactions between environmental conditions and regional functions. Such limitations become particularly pronounced under rare conditions, when traffic responses may differ substantially across regions with distinct functional attributes. Moreover, these approaches offer limited support for explicit semantic reasoning about how external conditions influence regional traffic dynamics, leaving contextual relationships insufficiently explored. Under the \textbf{imbalanced distribution} of observed traffic data, adverse scenarios are underrepresented, encouraging models to prioritize routine patterns during training. These models struggle to generalize to nonlinear traffic changes arising from complex external conditions.

Large language models (LLMs) offer a promising complement to conventional historical-data forecasting methods~\cite{liu2025timecma, zhang2025lsdm, xiao2025llm}. Trained on extensive textual corpora, LLMs possess broad world knowledge and can reason about contextual relationships. Given structured descriptions combining a region’s POI composition, weather conditions, and holiday status, LLMs can suggest plausible effects on human mobility, including responses that are sparsely represented in historical observations. These semantic priors may supplement limited empirical evidence and support generalization under rare or extreme conditions~\cite{brown2020language, kojima2022zeroshot}. The central methodological challenge lies in integrating semantic guidance from LLMs into quantitative spatiotemporal forecasting while preserving the numerical relationships learned from traffic observations.

In this paper, we aim to improve urban traffic flow prediction under adverse weather by combining semantic reasoning from LLMs with spatiotemporal graph learning. Achieving this objective presents three closely related technical challenges. \textbf{First}, structured semantic representations must capture interactions between external conditions and traffic dynamics. This requires transforming unstructured, heterogeneous contextual descriptions into features that characterize how adverse weather may affect regions with distinct functional compositions. \textbf{Second}, information from POI distributions, weather vectors, holiday labels, and traffic histories must be dynamically integrated, as the relevance and influence of these inputs vary across locations and over time. For example, heavy rainfall may affect travel demand differently in business and residential districts. \textbf{Third}, the contribution of semantic information must be carefully regulated, particularly when supervision for rare events is limited. Addressing this challenge requires a robust, interpretable correction mechanism that uses contextual semantic priors to calibrate backbone forecasts while preserving reliable numerical patterns, balancing predictive stability with adaptability to unusual conditions.

To address these challenges, we propose \textbf{Multi-STG}, a unified framework that integrates LLM semantic priors with a scalable graph Transformer backbone. First, we design an LLM-empowered semantic enhancement module to extract contextual semantic features. This module constructs structured prompts from regional information, including POI types and weather conditions, and queries a pretrained LLM to obtain semantic embeddings. Lightweight adapter networks then align these embeddings with the numerical feature space. Second, we introduce a multimodal attentional fusion module to dynamically integrate heterogeneous information. Conditioned on the current spatiotemporal state, this module adaptively weights traffic, semantic, and POI embeddings to prioritize the most relevant signals. Finally, we develop a semantic-enhanced prediction head with a gated residual correction mechanism to regulate the contribution of semantic information. The gate explicitly controls the extent of semantic correction, allowing contextual priors to refine backbone forecasts while preserving the numerical patterns learned from traffic observations.

Our contributions are summarized as follows:
\begin{itemize}
  \item We identify key limitations of existing traffic forecasting models under adverse weather, particularly sparse observations and limited semantic reasoning capabilities. We highlight the importance of integrating contextual information, including weather conditions and POI distributions, to capture complex nonlinear traffic variations in rare scenarios.
  
  \item We propose Multi-STG, a unified framework that integrates LLM-derived semantic priors with a graph backbone. The framework comprises an LLM-empowered semantic enhancement module to extract contextual semantic features, a multimodal attentional fusion module to adaptively weight heterogeneous inputs, and a semantic-enhanced prediction head for gated residual calibration.
  
  \item We conduct extensive experiments on datasets of private vehicle traffic from three cities. The results demonstrate that Multi-STG outperforms leading baselines, achieving an average MAE reduction of 11.82\% relative to the strongest baseline while maintaining stable performance across all three cities.
\end{itemize}


%% file: Sec2_relatedwork.tex
\section{Related Work}
\subsection{Urban Traffic Prediction}
Urban traffic prediction has been extensively studied through spatiotemporal modeling and multimodal fusion~\cite{gong2023empowering, gong2024kgda}. Early deep learning approaches primarily employed recurrent or convolutional architectures to capture temporal and spatial dependencies~\cite{zhang2019decomposition}. Graph neural networks have enabled more explicit modeling of road network topology through graph convolutions and dynamic graph learning~\cite{song2020stsgcn, zhang2023deep, yu2022network}. Attention mechanisms further capture long-range spatial interactions and provide flexible spatiotemporal receptive fields~\cite{zheng2020gman, xu2020sttn}. Hybrid architectures combining graph learning with recurrent networks have also been explored to model spatial dependencies that evolve with changing traffic conditions~\cite{li2023dgcrn, long2023practical}.

Beyond structural dependencies, recent studies have emphasized the importance of incorporating external conditions into traffic prediction. Models that integrate POI semantics, event information, holiday patterns, or regional functional attributes can better capture heterogeneous urban dynamics~\cite{zhang2026lg, zhang2022cityneuro, chai2024knowledge}. Weather conditions have received particular attention because adverse weather can induce substantial nonlinear changes in mobility patterns. Multimodal frameworks demonstrate the benefits of integrating meteorological observations, contextual information, and mobility features to improve forecasting robustness~\cite{koesdwiady2016improving, zhong2024weather}. Generative and probabilistic spatiotemporal models have also been developed to characterize predictive uncertainty under adverse weather conditions~\cite{wenDiffSTGProbabilisticSpatioTemporal2024, huUnifyingDiffusionModels2024, long2026dynamic, hui2023large, li2024mobile}.

Despite these advances, existing approaches primarily learn statistical relationships from historical observations and incorporate a limited range of modalities. Using semantic and commonsense knowledge to reason about traffic responses under rare or extreme conditions, particularly adverse weather, remains underexplored, motivating the integration of language models with broad contextual knowledge into traffic prediction.

\subsection{LLMs in Traffic Prediction}
Large language models (LLMs) offer rich semantic priors and contextual reasoning capabilities, motivating growing interest in urban computing applications. Recent studies have explored their potential for urban representation learning, multimodal reasoning, and mobility understanding. Some approaches align textual descriptions with mobility data to pretrain representations of urban regions or trajectories~\cite{yan2024urbanclip, feng2025agentmove}. Others use prompting or language-model-guided embedding refinement to incorporate contextual semantics into downstream mobility tasks~\cite{jin2025gurpp}.

Specialized LLMs have emerged to support urban analytics, mobility understanding, and traffic reasoning. Foundation models tailored to urban data demonstrate the value of incorporating textual and commonsense knowledge into urban planning and spatiotemporal perception~\cite{li2024urbangpt, feng2025citygpt}. In traffic modeling, recent studies have explored mobility analysis using LLMs, external-knowledge-informed traffic flow reasoning, and multimodal traffic generation~\cite{zhang2024trafficgpt, liu2025timecma}. Models focusing on weather and disaster contexts have also been proposed to support mobility forecasting under extreme conditions~\cite{shao2026weast}.

Existing traffic research using LLMs primarily focuses on semantic understanding and descriptive reasoning, with limited attention to directly improving spatiotemporal prediction. Few studies examine how to balance LLM-derived semantic knowledge with temporal patterns learned from traffic observations, particularly under adverse weather, when semantic priors may exert excessive influence and introduce prediction bias. To address this gap, our framework combines LLM-based semantic enhancement, selective multimodal fusion, and gated residual calibration to improve the robustness and interpretability of traffic forecasting under adverse weather conditions.

%% file: Sec3_preliminaries.tex
\section{Preliminaries}

\begin{figure*}[tb]
\vspace{-3mm}
\centering
\includegraphics[width=1\linewidth, trim=0 10 0 0, clip]{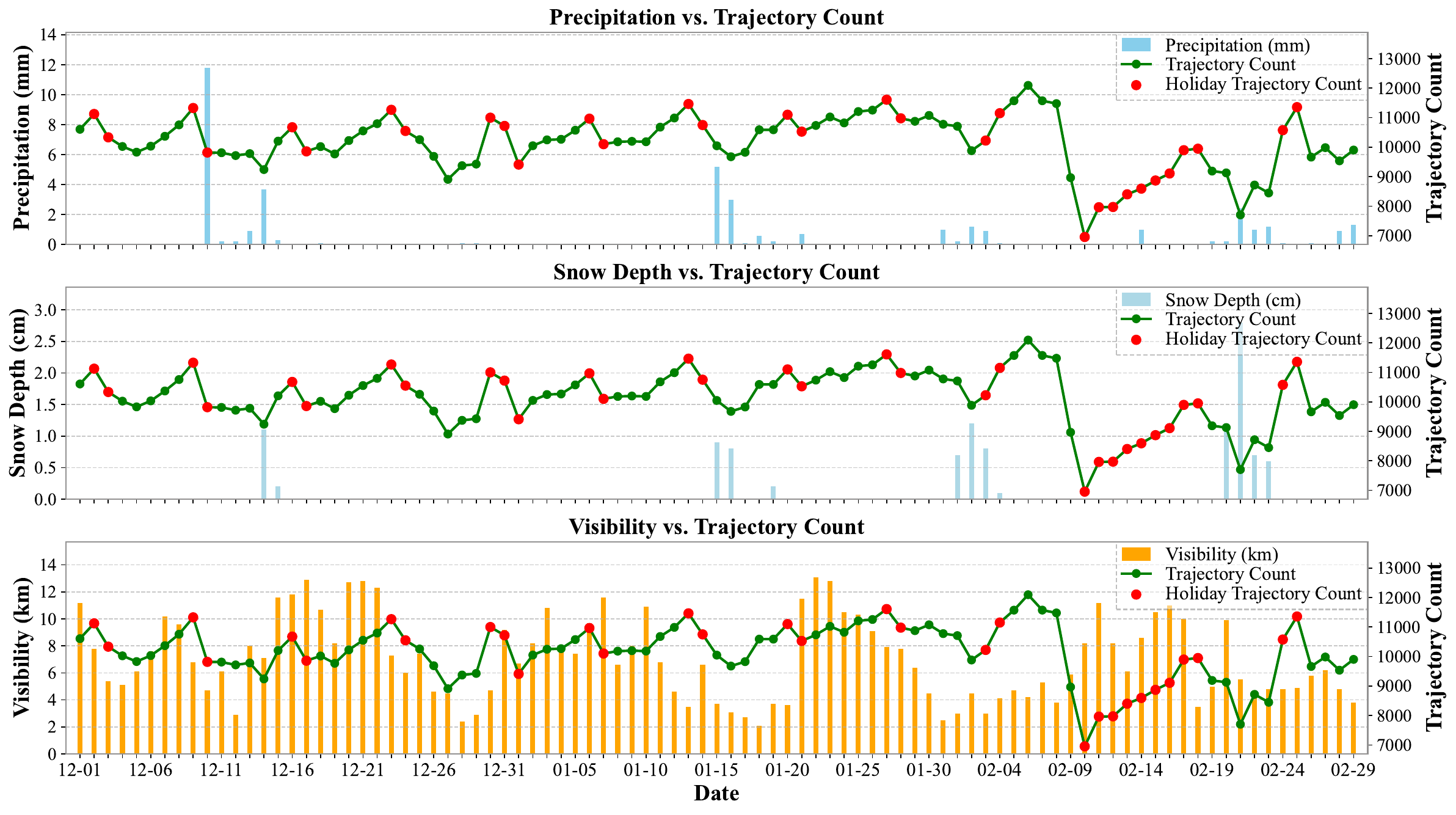}
\vspace{-3mm}
\caption{Statistical analysis of traffic trajectory volume, weather conditions, and holiday events in Xi'an, illustrating the correlation between traffic flow fluctuations and external environmental factors.}
\label{fig:weather_traffic_stats}
\vspace{-3mm}
\end{figure*}

\subsection{Urban Traffic Flow}
Urban traffic flow quantifies vehicle movement within a city and reflects travel demand, urban activity, and transportation system performance. Traffic patterns are shaped by urban structure, weather conditions, and social activities. Accurate forecasts support intelligent transportation systems by enabling congestion management, traffic signal optimization, and efficient road resource allocation. Reliable predictions also support individual route planning, commercial logistics, and urban infrastructure planning. Under adverse weather conditions, timely forecasts can help transportation authorities anticipate disruptions and maintain operational efficiency.

To represent urban traffic flow, a city is typically partitioned into mutually disjoint regions, denoted by \(\mathcal{R}=\{r_1,\dots,r_N\}\). These regions may be defined using geographic grids, administrative boundaries, or clustering methods that group areas with similar mobility characteristics. For each region \(r_i\), traffic flow during interval \(t\), denoted by \(x_i(t)\), is defined as the total number of trip departures and arrivals recorded within that region, corresponding to the sum of regional outflow and inflow. Aggregating these observations yields a citywide traffic vector \(\mathbf{x}(t)\in\mathbb{R}^N\). The sequence \(\{\mathbf{x}(1),\dots,\mathbf{x}(T)\}\) forms a multivariate time series. Traffic flow exhibits temporal dependencies, including recurring commuting peaks and demand changes during holidays, as well as spatial correlations among neighboring regions or regions with similar functions. These spatial relationships are commonly represented by an urban graph \(\mathcal{G}=(\mathcal{R},\mathbf{A})\), where \(\mathbf{A}\in\mathbb{R}^{N\times N}\) encodes geographic proximity, road connectivity, or mobility connections between regions. Under this formulation, traffic flow prediction aims to estimate \(\mathbf{x}(t+\tau)\) at a future horizon \(\tau\), conditioned on historical observations \(\mathbf{x}(1{:}t)\) and relevant external information, such as weather conditions, POI distributions, and holiday schedules. This formulation captures the spatiotemporal evolution of urban mobility while accounting for contextual influences.

\subsection{Weather-Traffic Sensitivity}
We use a large collection of private vehicle trip records obtained from a telematics service provider in China, covering 2022 to 2024. For this study, we select three metropolitan areas and combine trip records with meteorological observations and statutory holiday schedules. Trip departure and arrival locations and timestamps are used to reconstruct regional traffic flows.

Fig.~\ref{fig:weather_traffic_stats} presents Xi'an as a representative example to illustrate associations between traffic volume, weather conditions, and holidays. The weather analysis covers December 2023 to February 2024, while the forecasting experiments use contextual information throughout the full modeling period. We consider precipitation, snowfall, and visibility as key meteorological variables.

Traffic volume generally declines as weather conditions deteriorate. Average daily volume decreases from 10,080 on dry days to 9,856 under light precipitation of up to 1 mm and 9,331 under precipitation above 1 mm. However, the latter category includes only eight days, so interpret it cautiously. Each additional centimeter of snowfall is associated with an approximately 5.6\% reduction in daily volume. Lower visibility also coincides with reduced traffic, although this association cannot be separated from accompanying conditions such as heavy rain and fog.

Holiday periods exhibit distinct departures from routine traffic patterns. During the Spring Festival, traffic volume peaks before the holiday, reaches a minimum during the holiday, and rebounds afterward. These fluctuations suggest that holiday schedules provide relevant context beyond regular temporal patterns.

\begin{table}[tb]
    \centering
    \vspace{-3mm}
    \caption{Precipitation sensitivity by POI composition in Xi'an, measured as the percentage change in daily flow per millimeter after accounting for weekday and holiday effects. Low and High denote the bottom and top terciles of each POI category's regional share. Diff.\ denotes High minus Low.}
     \vspace{-3mm}
    \label{tab:poi_sensitivity}
    \footnotesize
    \begin{tabular}{lrrr}
        \toprule
        \textbf{POI Category} & \textbf{Low} & \textbf{High} & \textbf{Diff.} \\
        \midrule
        Commercial Services       & $-1.77$ & $-1.19$ & $+0.58$ \\
        Cultural \& Entertainment & $-1.55$ & $-1.64$ & $-0.08$ \\
        Tourism \& Travel         & $-1.19$ & $-1.80$ & $-0.61$ \\
        Transportation Facilities & $-1.19$ & $-2.22$ & $-1.04$ \\
        Residential \& Office     & $-1.97$ & $-1.13$ & $+0.84$ \\
        \bottomrule
    \end{tabular}
    \vspace{-3mm}
\end{table}

Weather associations also vary across regions. After accounting for weekday and holiday effects, 84\% of the 120 regions in Xi'an exhibit a negative association between precipitation and traffic volume. The estimated mean and median reductions are 1.4\% and 0.6\% per millimeter, respectively. Table~\ref{tab:poi_sensitivity} compares precipitation sensitivity across regions with low and high shares of different POI categories, defined by the bottom and top terciles of each category’s share. Transportation facilities show the largest contrast, while other categories exhibit smaller differences. Together, these observations motivate contextual modeling that accounts for weather conditions, holiday patterns, and regional functional composition alongside spatiotemporal dependencies.

\subsection{Problem Definition}
\label{sec:problem_definition}
We consider an urban road network partitioned into \(N\) regions. At each time step \(t\), the observed traffic state is represented by \(\boldsymbol{X}_t \in \mathbb{R}^{N \times C}\), where \(C=3\) corresponds to regional traffic flow, time of day, and day of the week. Regional traffic flow is calculated as the sum of the inflow and outflow events defined above. In addition, the model incorporates the following contextual information.

\textbf{Regional POI Distribution:} For each region \(i\), we aggregate POIs across five functional categories to construct a feature vector \(\boldsymbol{P}_i \in \mathbb{R}^{D_{\text{poi}}}\), where \(D_{\text{poi}}=5\). These categories comprise commercial services, culture and entertainment, tourism and travel, transportation facilities, and residential and office functions. The resulting vector captures the region's functional composition.

\textbf{External Environmental Information:} We represent weather conditions and calendar attributes, including holiday indicators and date types, as \(\boldsymbol{E}_t\).

\textbf{LLM Semantic Features:} We construct structured prompts from the regional POI distribution \(\boldsymbol{P}_i\) and environmental information \(\boldsymbol{E}_t\). A pretrained LLM processes these prompts to obtain a semantic representation \(\boldsymbol{S}_{i,t} \in \mathbb{R}^{D_{\text{sem}}}\) that characterizes potential contextual influences on traffic in region \(i\) at time \(t\).

The input for each region and time step combines \(\boldsymbol{X}_{i,t}\), \(\boldsymbol{P}_i\), and \(\boldsymbol{S}_{i,t}\). Given a historical window of length \(T\), the model learns a mapping \(f_{\boldsymbol{\Theta}}(\cdot)\) to predict traffic flows over the next \(T'\) time steps:
\begin{equation}
[\boldsymbol{X}'_{t+1}, \ldots, \boldsymbol{X}'_{t+T'}] =
f_{\boldsymbol{\Theta}}
\left(
[\boldsymbol{X}_{t-T+1}, \ldots, \boldsymbol{X}_t],
\boldsymbol{P},
\boldsymbol{S}_{t-T+1:t}
\right),
\end{equation}
where $\boldsymbol{P} \in \mathbb{R}^{N \times D_{{poi}}} $ contains the POI features of all regions, \(\boldsymbol{S}_{t-T+1:t}\) contains the corresponding semantic features within the historical window, and \(\boldsymbol{\Theta}\) denotes the model parameters. Each output \(\boldsymbol{X}'_{t+\tau} \in \mathbb{R}^{N \times 1}\) contains the predicted regional traffic flows at horizon \(\tau\).

%% file: Sec4_method.tex
\section{methodology}
\begin{figure*}[tb]
    \centering
    \vspace{-3mm}
    \includegraphics[width=\textwidth]{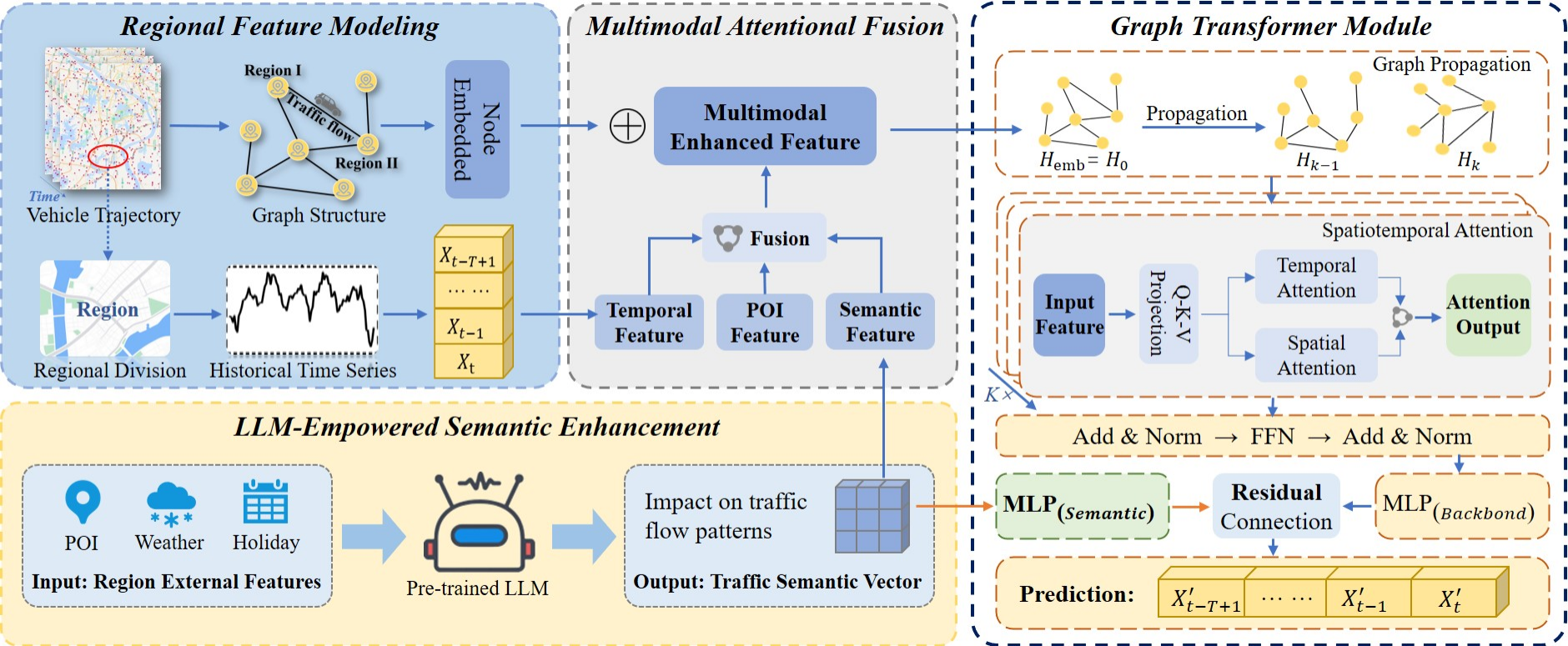}
    \vspace{-4mm}
    \caption{Overview of the proposed Multi-STG framework. The architecture consists of four key components: (1) Regional Feature Modeling for grid-based region partitioning and initial spatiotemporal encoding; (2) LLM-Empowered Semantic Enhancement for inferring traffic impacts from external contexts (weather, holidays); (3) Multimodal Attentional Fusion for dynamically integrating traffic, semantic, and POI features; and (4) a Graph Transformer Module that combines learned graph propagation with spatial and temporal attention, culminating in a semantic-enhanced residual prediction.}
    \label{fig:framework}
    \vspace{-6mm}
\end{figure*}

\subsection{Framework Overview}
In this section, we present Multi-STG, a framework for robust urban traffic flow prediction under adverse weather conditions. Multi-STG integrates semantic reasoning from large language models (LLMs) with spatiotemporal graph learning. As illustrated in Fig.~\ref{fig:framework}, the framework comprises four main modules:

\emph{Regional Feature Modeling:} This module processes raw trajectory data, partitions the city into a regular grid, and constructs initial spatiotemporal representations from historical regional traffic flows.

\emph{LLM-Empowered Semantic Enhancement:} This module uses a pretrained LLM to process regional context, including POI composition, weather conditions, and holiday information. The resulting semantic representations characterize potential nonlinear influences of external conditions on regional traffic dynamics.

\emph{Multimodal Attentional Fusion:} This module dynamically integrates static POI features, LLM semantic embeddings, and historical spatiotemporal representations. An attention mechanism with adaptive gating weights assigns each modality's contribution based on the current context.

\emph{Graph Transformer Module:} As the predictive backbone, this module propagates information over a learned adaptive graph to capture spatial dependencies, followed by spatiotemporal attention to model global spatial and temporal interactions. A semantic prediction head then applies gated residual corrections to refine the backbone forecasts while preserving numerical patterns learned from traffic observations.

\subsection{Regional Feature Modeling}

Regional Feature Modeling provides the foundation for our framework by transforming private vehicle trajectory records into structured regional representations. These representations capture traffic dynamics, including commuting patterns and weekday-weekend variations, and define the spatial units for subsequent semantic enhancement and graph learning. The process comprises two stages: dividing urban regions using a regular grid and modeling urban spatiotemporal dynamics.

\subsubsection{Urban Region Division}
We partition each city into a regular grid covering the geographic extent of the trajectory data. Specifically, we determine a bounding box for each city and divide the enclosed area into cells measuring \(2\,\mathrm{km}\times2\,\mathrm{km}\). All cells are retained as regions, including those with low traffic volumes, without filtering or merging based on traffic density. This procedure yields \(N=120\) regions for Xi'an, \(N=153\) for Shenyang, and \(N=30\) for Urumqi.

The grid partition offers three practical advantages over clustering methods. First, once the bounding box and grid alignment are fixed, the partition is deterministic and requires no model fitting. Second, fixed cell boundaries support consistent spatial aggregation as traffic volumes vary across seasons and cities. Third, uniform cell sizes provide a common spatial resolution that facilitates cross-city comparisons. A potential limitation is that individual cells may contain diverse land uses. We therefore explicitly incorporate each cell's POI composition, as described in Section~\ref{sec:problem_definition}, to represent regional functional characteristics.

\subsubsection{Urban Spatial-Temporal Modeling}
Urban traffic exhibits spatial dependencies among regions. Using the regional partition \(\mathcal{R}\), we represent the citywide traffic system as a directed graph \(\mathcal{G}=(\mathcal{V},\mathcal{E})\), where the nodes \(\mathcal{V}=\{v_1,\dots,v_N\}\) correspond to the regions in \(\mathcal{R}\). We learn the adjacency matrix from trainable node embeddings:
\begin{equation}
\mathbf{A}_t =
\mathrm{softmax}\left(
\mathrm{ReLU}\left(\mathbf{E}_t\mathbf{E}_t^{\top}\right)
\right),
\end{equation}
where \(\mathbf{E}_t\in\mathbb{R}^{N\times d}\) denotes the node embedding matrix, and softmax normalization is applied over each row. This formulation allows the model to learn spatial relationships jointly with forecasting.

To represent temporal dynamics, we aggregate trip events into hourly intervals. Let \(x_i(t)\) denote the traffic flow in region \(r_i\) during interval \(t\), defined as the total number of trip departures and arrivals recorded in that region. Observations over \(T\) intervals form a flow matrix \(\mathbf{X}^{\text{raw}}\in\mathbb{R}^{N\times T}\), with each row representing the temporal sequence of a region. We augment each flow observation with two temporal covariates, time of day and day of week, yielding an input tensor \(\mathbf{X}\in\mathbb{R}^{T\times N\times3}\). Together, the learned adjacency matrix and the input tensor provide a unified spatiotemporal representation for modeling dependencies across regions and over time.

\subsection{LLM-Empowered Semantic Enhancement}

Urban traffic patterns are shaped by road network structure, historical flow dynamics, and contextual factors such as weather conditions, holidays, and the spatial distribution of Points of Interest (POIs). These contextual factors influence both daily fluctuations and recurring demand patterns. For example, adverse weather can suppress mobility, while public holidays can alter travel demand within and between cities. 

\begin{figure*}[tb]
    \vspace{-3mm}
    \centering
    \includegraphics[width=0.9\textwidth,keepaspectratio]{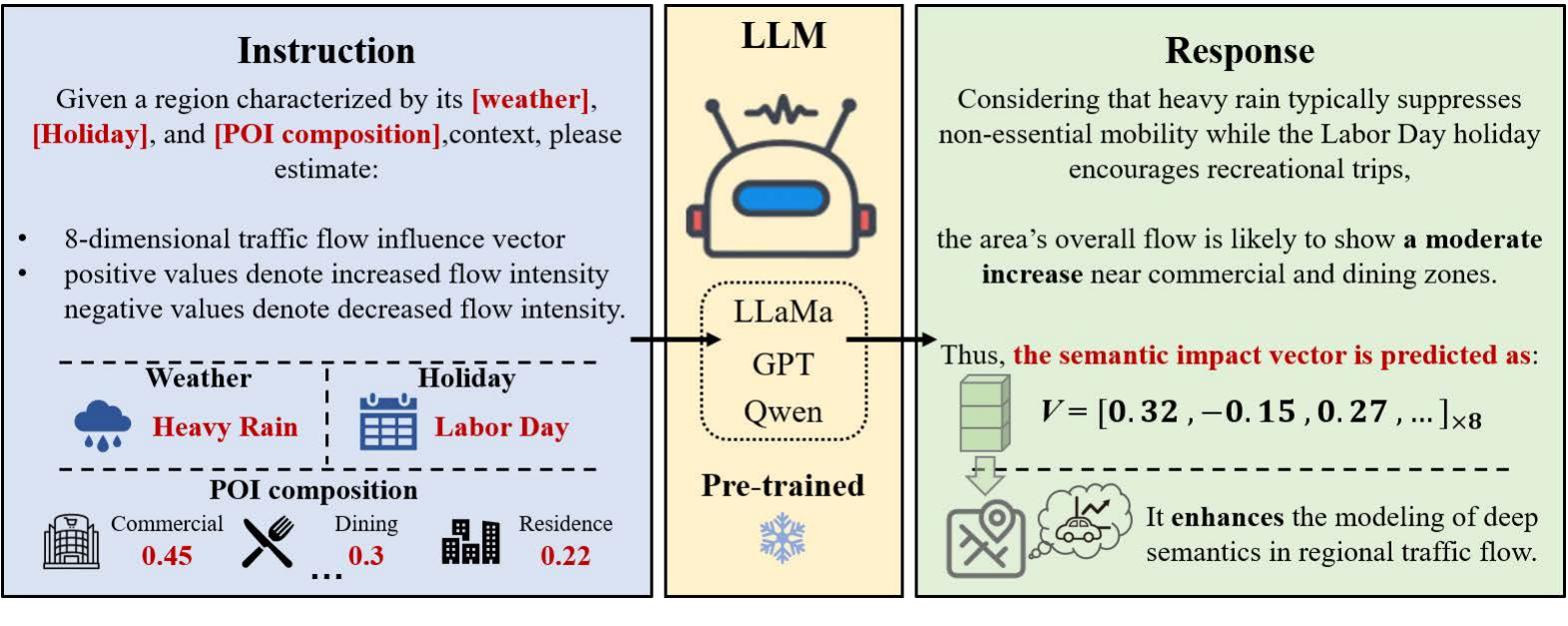}
    \vspace{-3mm}
    \caption{Example of inferring traffic flow influence using a pre-trained LLM.
The prompt is structured with task instruction, contextual factors (weather and holiday), and regional POI composition. The LLM generates a semantic reasoning response and outputs an 8-dimensional influence vector with entries in \([-1,1]\) indicating the intensity of traffic flow variation.}
    \label{fig:prompt}
    \vspace{-6mm}
\end{figure*}

Existing traffic prediction models commonly encode contextual variables using numerical embeddings. Although these representations provide useful information, they may inadequately capture interactions among environmental conditions and regional functions. Heavy rain or snow, for example, may affect traffic differently in commercial and residential areas, particularly during holidays. To better represent these interactions, we introduce semantic reasoning from a pretrained LLM. By drawing on contextual knowledge, the LLM provides semantic priors that describe potential traffic responses under different combinations of regional and environmental conditions.

For each region \(r_i\), we summarize the POI distribution as \(\mathbf{p}_i\in\mathbb{R}^{D_{\text{poi}}}\), where each dimension represents a functional category. At time step \(t\), weather features \(w_t\) and holiday labels \(h_t\) provide additional context. Together, these variables form the contextual feature set \(\mathcal{E}_{i,t}=\{\mathbf{p}_i,w_t,h_t\}\). A prompt construction function \(\Phi(\cdot)\) converts these structured features into a natural language description:
\begin{equation}
\text{Prompt}{i,t}=\Phi(\mathcal{E}{i,t}).
\end{equation}
The prompt describes the regional functional composition, weather conditions, and holiday status, and requests a structured assessment of their potential influence on traffic. This formulation enables the LLM to interpret the contextual features jointly.

As illustrated in Fig.~\ref{fig:prompt}, the pretrained LLM \(\mathcal{M}\) processes the prompt and produces a semantic feature vector:
\begin{equation}
\mathbf{s}_{i,t} = \mathcal{M}(\text{Prompt}{i,t}) \in\mathbb{R}^{D{\text{sem}}}.
\end{equation}
We set \(D_{\text{sem}}=8\), with dimensions corresponding to the morning peak, evening peak, daytime activity, nighttime activity, leisure demand, weather influence, holiday influence, and overall traffic level. This predefined structure organizes the semantic output into interpretable aspects of regional traffic dynamics. The resulting representation supplements the original contextual variables with semantic assessments of their combined influence. For example, heavy snowfall during a holiday may affect leisure demand and commuting activity differently across regions with distinct functional compositions. These semantic priors provide additional contextual guidance for subsequent spatiotemporal prediction.

\subsection{Multimodal Attentional Fusion}

To improve responsiveness to changes in external conditions, we introduce a dynamic fusion module that combines attention with adaptive gating. The module takes traffic features \(\mathbf{X}\), semantic features \(\mathbf{S}\) generated by the LLM, and POI features \(\mathbf{P}\) as inputs. Traffic features provide attention queries, semantic features provide keys, and POI features provide values. A complementary gating mechanism adjusts each modality's contribution at every time step and region. Together, we capture cross-region dependencies and adapt modality contributions to the current context.

The input features are represented as \(\mathbf{X}\in\mathbb{R}^{T\times N\times D_x}\), \(\mathbf{S}\in\mathbb{R}^{T\times N\times D_{\text{sem}}}\), and \(\mathbf{P}\in\mathbb{R}^{N\times D_{\text{poi}}}\), where \(D_x=C=3\). Using column vectors, we construct the attention queries, keys, and values as
\begin{equation}
\mathbf{q}_{t,n}=\mathbf{W}_Q\mathbf{x}_{t,n},\qquad
\mathbf{k}_{t,m}=\mathbf{W}_K\mathbf{s}_{t,m},\qquad
\mathbf{v}_m=\mathbf{W}_V\mathbf{p}_m,
\end{equation}
where the learned projection matrices map all three representations into a common space of dimension \(d\). The attention weights are computed as
\begin{equation}
\alpha_{t,n,m} = \operatorname{softmax}_m
\left(
\frac{\mathbf{q}_{t,n}^{\top}\mathbf{k}_{t,m}}{\sqrt{d}}
\right).
\end{equation}
The resulting attention representation aggregates POI information across regions:
\begin{equation}
\mathbf{a}_{t,n} = \sum_{m=1}^{N}\alpha_{t,n,m}\mathbf{v}_m.
\end{equation}
This formulation allows each region to retrieve functional information according to the compatibility between the local traffic state and the semantic context of other regions.

To regulate modality contributions, we additionally project the traffic, semantic, and POI features into aligned representations \(\widetilde{\mathbf{x}}_{t,n}\), \(\widetilde{\mathbf{s}}_{t,n}\), and \(\widetilde{\mathbf{p}}_n\), each belonging to \(\mathbb{R}^{d}\). These representations are concatenated and passed through a lightweight mapping, followed by softmax normalization:
\begin{equation}
\mathbf{g}_{t,n} = \operatorname{softmax}
\left(
\operatorname{MLP}
\left(
[\widetilde{\mathbf{x}}_{t,n}
\Vert\widetilde{\mathbf{s}}_{t,n}
\Vert\widetilde{\mathbf{p}}_n]
\right)
\right)
\in\mathbb{R}^{3},
\end{equation}
where \(\Vert\) denotes concatenation. The three components correspond to the traffic, semantic, and POI modalities, respectively. The gated representation is then obtained as
\begin{equation}
\mathbf{g}^{\text{fuse}}_{t,n} = g_{t,n}^{(1)}\widetilde{\mathbf{x}}_{t,n}
+ g_{t,n}^{(2)}\widetilde{\mathbf{s}}_{t,n} +
g_{t,n}^{(3)}\widetilde{\mathbf{p}}_n.
\end{equation}
The gating weights vary across time and regions, allowing the model to adjust modality contributions under different conditions, including adverse weather.

Finally, we combine the gated representation with the attention output through residual fusion:
\begin{equation}
\mathbf{F}_{t,n} = \mathbf{g}^{\text{fuse}}_{t,n}
+ \gamma \cdot \mathbf{a}_{t,n},
\end{equation}
where \(\gamma\) is a learnable scaling parameter. This design combines direct modality weighting with semantic guidance for aggregating regional functional information, producing the fused representation for the downstream predictor.

\subsection{Graph Transformer Module}

\begin{figure}[tb]
    \centering
    \vspace{-3mm}
    \includegraphics[width=0.6\linewidth]{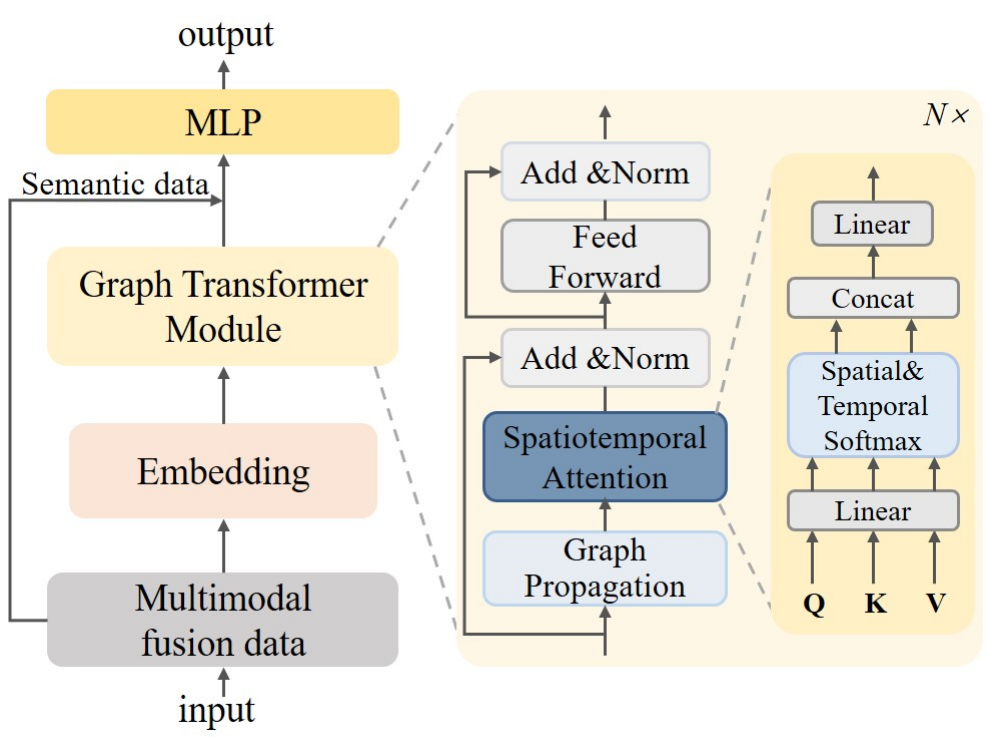}
    \vspace{-3mm}
    \caption{Graph Transformer architecture with adaptive graph propagation and gated fusion of representations at zero and one hop, spatial and temporal attention, and a semantic residual prediction head for contextual calibration.}
    \label{fig:graph_transformer}
    \vspace{-3mm}
\end{figure}

Urban traffic systems exhibit complex spatiotemporal dependencies arising from interactions among regions and external disturbances. To model these dependencies, we introduce the Graph Transformer Module shown in Fig.~\ref{fig:graph_transformer}. Given the fused input tensor $\mathbf{H}^{(0)}\in\mathbb{R}^{T\times N\times D}$, the module performs gated graph propagation, spatial and temporal attention, and semantic residual calibration.

The module first propagates information through the learned adjacency matrix $\mathbf{A}_t$. At each time step $t$, the original representation $\mathbf{H}^{(0)}_t\in\mathbb{R}^{N\times D}$ preserves the features of individual regions, while one propagation step aggregates information according to the learned graph:
\begin{equation}
\mathbf{H}^{(1)}_t
=
\mathbf{A}_t\mathbf{H}^{(0)}_t.
\end{equation}
A learnable gate balances the original and propagated representations:
\begin{equation}
\mathbf{H}_t
=
(1-\boldsymbol{\beta})\odot\mathbf{H}^{(0)}_t
+
\boldsymbol{\beta}\odot\mathbf{H}^{(1)}_t,
\end{equation}
where $\boldsymbol{\beta}$ contains gating values in $[0,1]$, broadcast as needed, and $\odot$ denotes elementwise multiplication. This operation preserves regional information while incorporating dependencies encoded by the learned adjacency. The connected regions need not be geographically adjacent.

Spatial and temporal attention subsequently capture interactions across regions and historical observations. The propagated features are projected into queries, keys, and values:
\begin{equation}
\mathbf{Q}=\mathbf{H}\mathbf{W}_Q,\qquad
\mathbf{K}=\mathbf{H}\mathbf{W}_K,\qquad
\mathbf{V}=\mathbf{H}\mathbf{W}_V,
\end{equation}
where $\mathbf{W}_Q$, $\mathbf{W}_K$, and $\mathbf{W}_V$ are learned projection matrices applied across all regions and time steps.

Spatial attention aggregates information across regions at each time step. The conventional attention formulation is
\begin{equation}
\mathbf{O}^{\mathrm{spatial}}_t
=
\operatorname{softmax}
\left(
\frac{\mathbf{Q}_t\mathbf{K}_t^{\top}}{\sqrt{d_k}}
\right)
\mathbf{V}_t,
\end{equation}
where $\mathbf{Q}_t,\mathbf{K}_t\in\mathbb{R}^{N\times d_k}$, and softmax normalization is applied over the source regions. Each target region aggregates information according to the compatibility between the query and the keys of other regions, allowing interactions beyond geographic proximity to contribute to prediction.

Temporal attention captures dependencies across historical time steps within each region:
\begin{equation}
\mathbf{O}^{\mathrm{temporal}}_n
=
\operatorname{softmax}
\left(
\frac{\mathbf{Q}^{(n)}
(\mathbf{K}^{(n)})^{\top}}{\sqrt{d_k}}
\right)
\mathbf{V}^{(n)},
\end{equation}
where $\mathbf{Q}^{(n)},\mathbf{K}^{(n)}\in\mathbb{R}^{T\times d_k}$ contain the query and key representations for region $n$ over the historical window. This operation allows different observations to contribute according to their relevance, supporting the modeling of recurring patterns and transient fluctuations.

The equations above describe the conventional softmax attention formulation. For computational efficiency, the implementation uses linearized attention, which replaces the softmax kernel with a feature-map approximation and aggregates key and value information before applying the queries. For fixed feature dimensions, this changes the computational dependence on the number of regions from quadratic to linear. The temporal counterpart similarly avoids constructing a full attention matrix over time. Although the approximation may affect attention estimates, no noticeable loss in prediction accuracy was observed for the shallow attention layers and city graphs considered in this study. A systematic assessment of the relationship between accuracy and efficiency remains a direction for future work.

The spatial and temporal outputs are concatenated and projected into a unified representation $\mathbf{Z}$. This representation combines interactions across regions with temporal dependencies within each region and provides the input to the forecasting head.

Finally, a semantic prediction head calibrates the backbone forecasts using contextual information from the LLM. Let
\begin{equation}
\mathbf{B}
=
f_{\mathrm{backbone}}(\mathbf{Z})
\in\mathbb{R}^{T'\times N\times1}
\end{equation}
denote the backbone predictions for the next $T'$ time steps. We broadcast the semantic features $\mathbf{S}_t$ across the prediction horizon to obtain $\overline{\mathbf{S}}_t$. A lightweight MLP processes the concatenated backbone predictions and semantic features to produce a residual correction:
\begin{equation}
\Delta\mathbf{X}
=
f_{\mathrm{sem}}
\left(
[\mathbf{B}\Vert\overline{\mathbf{S}}_t]
\right),
\end{equation}
where $\Vert$ denotes concatenation along the feature dimension, and $\Delta\mathbf{X}$ has the same dimensions as $\mathbf{B}$. The final predictions are
\begin{equation}
\mathbf{X}'_{t+1:t+T'}
=
\mathbf{B}
+
\lambda\,\Delta\mathbf{X},
\end{equation}
where $\lambda$ is a learnable residual weight controlling the contribution of semantic correction. This formulation allows contextual priors to refine the backbone forecasts while retaining an explicit prediction component learned from traffic observations.


%% file: Sec5_evaluation.tex
\section{Experiments}
In this section, we evaluate Multi-STG using private vehicle trajectory data and compare performance against established forecasting baselines. We also conduct ablation studies to assess individual components, experiments with varying amounts of training data to examine the effect of data availability, evaluations across cities to assess transferability, and hyperparameter sensitivity analyses to examine model stability.

\subsection{Datasets}

\begin{table*}[tb]
    \centering
    \footnotesize
    \caption{Statistics of the private vehicle trip dataset. \#Trips denotes deduplicated trips with at least one endpoint within the city and modeling period. \#Events counts eligible departures and arrivals, with two events when both endpoints qualify. \#Vehicles denotes unique vehicles in the selected trips.}
    \vspace{-3mm}
    \label{tab:dataset_statistics}
    \resizebox{1.0\linewidth}{!}{
    \begin{tabular}{@{}lrrrccccr@{}}
        \toprule
        \textbf{City}
        & \textbf{\#Trips}
        & \textbf{\#Vehicles}
        & \textbf{\#Events}
        & \textbf{Modeling Period}
        & \textbf{Longitude Range}
        & \textbf{Latitude Range}
        & \textbf{Grid}
        & \textbf{\#Regions} \\
        \midrule
        Xi'an
        & 1,024,825 & 18,695 & 1,743,732
        & 2023/12/01--2024/05/24
        & $108.85$--$109.05^{\circ}\mathrm{E}$
        & $34.15$--$34.35^{\circ}\mathrm{N}$
        & $12\times10$ & 120 \\
        
        Shenyang
        & 1,294,948 & 13,803 & 2,305,984
        & 2023/12/01--2024/05/24
        & $123.32$--$123.52^{\circ}\mathrm{E}$
        & $41.64$--$41.94^{\circ}\mathrm{N}$
        & $17\times9$ & 153 \\
        
        Urumqi
        & 519,226 & 6,429 & 767,107
        & 2023/12/01--2024/05/24
        & $87.55$--$87.65^{\circ}\mathrm{E}$
        & $43.75$--$43.85^{\circ}\mathrm{N}$
        & $6\times5$ & 30 \\
        \bottomrule
    \end{tabular}
    }
    \vspace{-3mm}
\end{table*}

\begin{table}[tb]
    \centering
    \caption{Attributes of the private vehicle trip dataset.}
    \vspace{-3mm}
    \label{tab:dataset_attributes}
    \footnotesize
    \resizebox{0.9\linewidth}{!}{
    \begin{tabular}{
        >{\centering\arraybackslash}p{0.16\columnwidth}
        >{\raggedright\arraybackslash}p{0.30\columnwidth}
        >{\centering\arraybackslash}p{0.16\columnwidth}
        >{\raggedright\arraybackslash}p{0.30\columnwidth}
    }
        \toprule
        \textbf{Attribute} & \textbf{Description}
        & \textbf{Attribute} & \textbf{Description} \\
        \midrule
        RecID & Unique trip record identifier
        & StartLon & Longitude of the trip origin \\

        ObjectID & Unique vehicle identifier
        & StartLat & Latitude of the trip origin \\

        StartTime & Trip start timestamp
        & StopLon & Longitude of the trip destination \\

        StopTime & Trip end timestamp
        & StopLat & Latitude of the trip destination \\
        \bottomrule
    \end{tabular}
    }
    \vspace{-3mm}
\end{table}

\begin{table*}[tb]
\centering
\caption{POI category aggregation and functional rationale.}
\vspace{-3mm}
\label{tab:poi_categories}
\footnotesize
\resizebox{0.9\linewidth}{!}{
\begin{tabular}{@{}p{0.19\linewidth}p{0.30\linewidth}p{0.45\linewidth}@{}}
\toprule
\textbf{Merged Category}
& \textbf{Original Categories Included}
& \textbf{Rationale for Merging} \\
\midrule
\textbf{Commercial Services}
& Dining, Shopping, Lifestyle Services, Companies, Finance, Automotive, Healthcare Facilities
& Groups destinations associated with commercial and service activities, encompassing routine visits, discretionary travel, and essential medical trips. \\
\textbf{Cultural \& Entertainment}
& Leisure, Education, Sports, Culture
& Combines destinations for educational, cultural, sporting, and leisure activities, with travel demand shaped by activity schedules, weather conditions, and holidays. \\
\textbf{Tourism \& Travel}
& Scenic Spots, Hotels, Lodging
& Groups tourist attractions and accommodation facilities associated with visitor activity, holiday travel, and movement between regions. \\
\textbf{Transportation Facilities}
& Transportation Hubs, including Stations, Airports, and Ports
& Represents mobility terminals connecting local and regional travel, with demand influenced by transport schedules and holiday travel patterns. \\
\textbf{Residential \& Office}
& Residential, Business, Office Areas
& Captures residential origins and workplace destinations that structure routine commuting and recurring travel between regions. \\
\bottomrule
\end{tabular}
}
\vspace{-3mm}
\end{table*}

We use a private vehicle trajectory dataset~\cite{li2019driving} collected in three Chinese cities: Xi'an, Shenyang, and Urumqi. The six monthly files contain 5,144,177 deduplicated trips. Among these, 2,838,999 trips fall within the modeling window and have at least one endpoint inside the corresponding city boundary. These trips yield 4,816,823 inflow and outflow events, comprising 1,743,732 events in Xi'an, 2,305,984 in Shenyang, and 767,107 in Urumqi. The modeling window extends from 1 December 2023 through 24 May 2024, covering 4,224 hourly intervals. Each city is partitioned into a regular grid of \(2\,\mathrm{km}\times2\,\mathrm{km}\) cells, yielding 120, 153, and 30 regions, respectively. Regional traffic flow is defined as the total number of trip departures and arrivals recorded within each region during an hourly interval. Table~\ref{tab:dataset_statistics} summarizes the dataset statistics.

Before training, we preprocess the data by removing duplicates, filtering anomalous records, aggregating hourly, and denoising sparse regional time series by temporal period. Denoising statistics are estimated exclusively from the training segment, and the same preprocessing pipeline is applied to all models. As summarized in Table~\ref{tab:dataset_attributes}, the raw attributes include vehicle and trip identifiers, departure and arrival timestamps, and origin and destination coordinates.

To represent regional functionality, we construct functional profiles using Points of Interest (POIs) obtained from open map APIs. Each POI is assigned to a region according to the geographic coordinates. To reduce sparsity while preserving functional distinctions, we consolidate the 14 original categories into five classes: Commercial Services, Cultural \& Entertainment, Tourism \& Travel, Transportation Facilities, and Residential \& Office. The category mapping is detailed in Table~\ref{tab:poi_categories}. For each region, we calculate the proportion of POIs in each class, producing a five-dimensional feature vector. This vector characterizes regional functional composition and provides static semantic context for modeling traffic patterns across heterogeneous urban regions.

Data providers and participating users authorized data access and use. All records were anonymized before acquisition, and traffic data used for modeling were aggregated by region to protect personal information.

\subsection{Metrics and Baselines}

We evaluate forecasting performance using MAE, RMSE, WAPE, and $R^2$. MAE and RMSE are reported in flow units, with RMSE assigning greater weight to large errors. WAPE measures aggregate relative error and is defined as $\mathrm{WAPE}= \frac{\sum_j|\hat{y}_j-y_j|}{\sum_j|y_j|}\times100\%$, where $j$ indexes the evaluated prediction points. $R^2$ evaluates predictive fit relative to a baseline that always predicts the mean observed flow. We compute all metrics globally by pooling predictions across test windows, forecast horizons, and regions. All models use the same preprocessed series for training and evaluation.

We compare Multi-STG with eight representative forecasting baselines:

$\bullet$ \textbf{LSTM}~\cite{hochreiter1997lstm}. Uses gated recurrence to capture temporal dependencies. The standard architecture provides a sequential baseline without explicitly modeling graph relationships among regions.

$\bullet$ \textbf{STGCN}~\cite{yu2018stgcn}. Combines graph convolutions with gated temporal convolutions to jointly capture spatial relationships and temporal dynamics.

$\bullet$ \textbf{GWaveNet}~\cite{2019graphWaveNet}. Combines adaptive graph learning with dilated causal convolutions to capture latent spatial connectivity and temporal dependencies over extended historical windows.

$\bullet$ \textbf{DSTAGNN}~\cite{2022lanDSTAGNN}. Models dynamic spatial dependencies and temporal patterns through learned graph relationships, spatial attention, and gated temporal convolutions.

$\bullet$ \textbf{TACTiS}~\cite{drouin2022tactis}. Uses a Transformer architecture and an attentional copula decoder to estimate joint predictive distributions for multivariate time series.

$\bullet$ \textbf{USTD}~\cite{huUnifyingDiffusionModels2024}. Provides a unified diffusion framework with a shared spatiotemporal encoder and specialized attention denoisers for forecasting and spatial interpolation.

$\bullet$ \textbf{DiffSTG}~\cite{wenDiffSTGProbabilisticSpatioTemporal2024}. Extends denoising diffusion probabilistic modeling to spatiotemporal graphs, enabling nonautoregressive forecasting and predictive uncertainty estimation.

$\bullet$ \textbf{SARAF}~\cite{zhou2026saraf}. Augments forecasting with historical retrieval, balancing temporal relevance and retrieval diversity and adjusting aggregation according to series stationarity.

\subsection{Experiment Settings}
We organize traffic observations, POI distributions, and weather data by region and align temporal observations to hourly intervals. The LLM-empowered Semantic Enhancement module derives semantic features from regional contextual information. The grid partitions yield graphs with 120, 153, and 30 nodes for Xi'an, Shenyang, and Urumqi, respectively. Each sample contains 24 hours of historical observations and 12 hours of target flows. Of the 4,189 candidate sliding windows, 2,932 are used for training, 396 for validation, and 839 for testing (approximately 7:1:2). To avoid target-time overlap, we discard 11 windows at the training/validation boundary and 11 at the validation/test boundary, leaving 4,167 used windows; historical input windows may overlap across splits, whereas target intervals are disjoint. We standardize the flow channel using statistics computed exclusively from the training set and transform predictions back to the original scale for evaluation.

The Graph Transformer backbone uses a model dimension of 152, four attention heads, and three layers, with a dropout rate of 0.1. The Multimodal Attentional Fusion module uses two attention heads, a hidden dimension of 64, and softmax gating to regulate modality contributions. We train the model using Adam with a learning rate of \(1\times10^{-3}\) and a batch size of 32. Training runs for up to 100 epochs, with an early stopping patience of 20 epochs and an \(\ell_1\) loss computed on standardized flow values. Unless otherwise specified, all experiments follow the same training configuration and evaluation protocol.

\subsection{Overall Performance Evaluation}

\begin{table*}[tb]
    \centering
    \vspace{-3mm}
    \caption{Forecasting performance across three cities. Best and second best results are shown in bold and underlined, respectively. Improv.\ reports percentage reductions in MAE, RMSE, and WAPE and absolute gains in $R^2$ relative to the strongest baseline for each city and metric.}
    \vspace{-3mm}
    \label{tab:baseline}
    \footnotesize
    \setlength{\tabcolsep}{3pt}
    \renewcommand{\arraystretch}{1.1}
    \resizebox{1.0\linewidth}{!}{
    \begin{tabular}{@{}l*{12}{c}@{}}
        \toprule
        \multirow{2}{*}{\textbf{Model}}
        & \multicolumn{4}{c}{\textbf{Xi'an}}
        & \multicolumn{4}{c}{\textbf{Shenyang}}
        & \multicolumn{4}{c}{\textbf{Urumqi}} \\
        \cmidrule(lr){2-5}
        \cmidrule(lr){6-9}
        \cmidrule(lr){10-13}
        & \textbf{MAE} & \textbf{RMSE} & \textbf{WAPE (\%)} & \textbf{R$^2$}
        & \textbf{MAE} & \textbf{RMSE} & \textbf{WAPE (\%)} & \textbf{R$^2$}
        & \textbf{MAE} & \textbf{RMSE} & \textbf{WAPE (\%)} & \textbf{R$^2$} \\
        \midrule
        DSTAGNN
        & \underline{0.8386} & \underline{1.3001} & \underline{25.58} & \underline{0.8725}
        & 0.8962 & \underline{1.4237} & 25.45 & \underline{0.9079}
        & 1.2280 & 1.9650 & 20.35 & 0.9277 \\

        STGCN
        & 0.8425 & 1.3407 & 25.70 & 0.8644
        & \underline{0.8936} & 1.4763 & \underline{25.37} & 0.9009
        & \underline{1.1732} & 1.9935 & \underline{19.44} & 0.9256 \\

        GWaveNet
        & 0.9131 & 1.4512 & 27.85 & 0.8411
        & 0.9498 & 1.5739 & 26.97 & 0.8874
        & 1.2415 & 2.0970 & 20.57 & 0.9176 \\

        SARAF
        & 0.8798 & 1.3426 & 26.84 & 0.8640
        & 0.9013 & 1.4336 & 25.59 & 0.9066
        & 1.1827 & \underline{1.9098} & 19.60 & \underline{0.9317} \\

        LSTM
        & 0.8936 & 1.4197 & 27.26 & 0.8479
        & 0.9324 & 1.5417 & 26.47 & 0.8920
        & 1.2011 & 2.0200 & 19.90 & 0.9236 \\

        TACTiS
        & 0.9948 & 1.5300 & 30.34 & 0.8234
        & 1.0760 & 1.7342 & 30.55 & 0.8633
        & 1.6163 & 2.6308 & 26.78 & 0.8704 \\

        USTD
        & 1.1301 & 1.7630 & 34.47 & 0.7655
        & 1.1893 & 1.9407 & 33.77 & 0.8288
        & 1.6421 & 2.7619 & 27.21 & 0.8571 \\

        DiffSTG
        & 1.6902 & 2.3281 & 51.55 & 0.5910
        & 1.8056 & 2.7049 & 51.27 & 0.6675
        & 1.9059 & 3.1326 & 31.58 & 0.8162 \\
        \midrule

        \textbf{Multi-STG}
        & \textbf{0.7556} & \textbf{1.1982} & \textbf{23.05} & \textbf{0.8917}
        & \textbf{0.7826} & \textbf{1.2836} & \textbf{22.22} & \textbf{0.9251}
        & \textbf{1.0189} & \textbf{1.6860} & \textbf{16.88} & \textbf{0.9468} \\
        \midrule

        \textbf{Improv.}
        & 9.90\% & 7.84\% & 9.89\% & $+0.0192$
        & 12.42\% & 9.84\% & 12.42\% & $+0.0172$
        & 13.15\% & 11.72\% & 13.17\% & $+0.0151$ \\
        \bottomrule
    \end{tabular}
    }
    \vspace{-6mm}
\end{table*}

Table~\ref{tab:baseline} compares Multi-STG with eight forecasting baselines across Xi'an, Shenyang, and Urumqi. The best and second-best results are shown in bold and underlined, respectively. Multi-STG achieves the best performance across all 12 metric-city combinations. Relative to the strongest baseline for each city and metric, Multi-STG reduces MAE by 9.90\% to 13.15\%, averaging 11.82\%, RMSE by 7.84\% to 11.72\%, and WAPE by 9.89\% to 13.17\%. The corresponding \(R^2\) values are 0.8917, 0.9251, and 0.9468 for Xi'an, Shenyang, and Urumqi, respectively.

DSTAGNN, STGCN, and SARAF provide the strongest baseline results in different settings. DSTAGNN leads across all four metrics in Xi'an and achieves the best baseline RMSE and \(R^2\) in Shenyang. STGCN achieves the lowest baseline MAE and WAPE in Shenyang and Urumqi, while SARAF performs best among the baselines on RMSE and \(R^2\) in Urumqi. LSTM remains competitive on selected metrics but does not match the strongest baseline results. The probabilistic models DiffSTG, USTD, and TACTiS produce higher point forecast errors in this evaluation. These results characterize the accuracy of their point predictions, while the quality of their predictive distributions requires separate assessment.

The consistent gains across all three cities demonstrate Multi-STG's overall effectiveness. The framework combines adaptive graph propagation, spatial and temporal attention, multimodal fusion, and contextual semantic information to model regional traffic dynamics. The results support this integrated design for urban traffic prediction across heterogeneous city settings.

\subsection{Ablation Studies}

\begin{table*}[tb]
    \centering
    \vspace{-3mm}
    \caption{Ablation results across three cities. STG (w/o Contextual Features), A-STG (w/o LLM-Empowered Semantic Enhancement), B-STG (w/o Multimodal Attentional Fusion), C-STG (w/o Semantic Enhanced Prediction Head), Multi-STG (Full Model).}
    \vspace{-3mm}
    \label{tab:ablation_results}
    \footnotesize
    \setlength{\tabcolsep}{3pt}
    \renewcommand{\arraystretch}{1.15}
    \resizebox{1.0\linewidth}{!}{
    \begin{tabular}{@{}l*{12}{c}@{}}
        \toprule
        \multirow{2}{*}{\textbf{Model}}
        & \multicolumn{4}{c}{\textbf{Xi'an}}
        & \multicolumn{4}{c}{\textbf{Shenyang}}
        & \multicolumn{4}{c}{\textbf{Urumqi}} \\
        \cmidrule(lr){2-5}
        \cmidrule(lr){6-9}
        \cmidrule(lr){10-13}
        & \textbf{MAE} & \textbf{RMSE} & \textbf{WAPE (\%)} & \textbf{R$^2$}
        & \textbf{MAE} & \textbf{RMSE} & \textbf{WAPE (\%)} & \textbf{R$^2$}
        & \textbf{MAE} & \textbf{RMSE} & \textbf{WAPE (\%)} & \textbf{R$^2$} \\
        \midrule

        STG
        & 0.8319 & 1.2952 & 25.38 & 0.8735
        & 0.8621 & 1.3864 & 24.48 & 0.9126
        & 1.1305 & 1.8247 & 18.73 & 0.9377 \\

        A-STG
        & 0.8124 & 1.2735 & 24.79 & 0.8777
        & 0.8402 & 1.3643 & 23.85 & 0.9153
        & 1.1048 & 1.7866 & 18.30 & 0.9402 \\

        B-STG
        & 0.7972 & 1.2528 & 24.31 & 0.8817
        & 0.8247 & 1.3416 & 23.41 & 0.9182
        & 1.0814 & 1.7563 & 17.91 & 0.9422 \\

        C-STG
        & {0.7815} & {1.2307} & {23.83} & {0.8857}
        & {0.8081} & {1.3188} & {22.94} & {0.9209}
        & {1.0566} & {1.7275} & {17.50} & {0.9442} \\
        \midrule

        \textbf{Multi-STG}
        & \textbf{0.7556} & \textbf{1.1982} & \textbf{23.05} & \textbf{0.8917}
        & \textbf{0.7826} & \textbf{1.2836} & \textbf{22.22} & \textbf{0.9251}
        & \textbf{1.0189} & \textbf{1.6860} & \textbf{16.88} & \textbf{0.9468} \\
        \bottomrule
    \end{tabular}
    }
    \vspace{-3mm}
\end{table*}

To assess the contributions of individual components and the effectiveness of their integration, we evaluate ablation variants of Multi-STG on the three city datasets under identical experimental settings. The variants are defined as follows:

\(\bullet\) \textbf{STG (w/o Contextual Features):} Uses the spatiotemporal forecasting backbone with traffic sequences alone, excluding external contextual and environmental information.

\(\bullet\) \textbf{A-STG (w/o LLM-Empowered Semantic Enhancement):} Incorporates POI distributions, weather conditions, and holiday indicators through direct feature encoding, without LLM-based semantic reasoning. This variant evaluates the additional contribution of semantic enhancement beyond directly encoded context.

\(\bullet\) \textbf{B-STG (w/o Multimodal Attentional Fusion):} Replaces the Multimodal Attentional Fusion module with simple feature concatenation to assess the contribution of adaptive modality weighting and selection.

\(\bullet\) \textbf{C-STG (w/o Semantic Enhanced Prediction Head):} Removes the semantic prediction head while retaining all other components, allowing the contribution of residual calibration to be evaluated.

\(\bullet\) \textbf{Multi-STG (Full Model):} Integrates the Graph Transformer Module, LLM-empowered semantic enhancement, Multimodal Attentional Fusion, and the Semantic Enhanced Prediction Head.

Table~\ref{tab:ablation_results} summarizes the results. MAE, RMSE, and WAPE decrease consistently from STG through A-STG, B-STG, and C-STG to Multi-STG, while \(R^2\) increases in the same sequence. The full model performs best across all 12 combinations of metrics and cities.

Using the full model as the reference, the MAE increases for C-STG are 3.43\%, 3.26\%, and 3.70\% in Xi'an, Shenyang, and Urumqi, respectively. The corresponding increases are 5.51\%, 5.38\%, and 6.13\% for B-STG, 7.52\%, 7.36\%, and 8.43\% for A-STG, and 10.10\%, 10.16\%, and 10.95\% for STG.

These comparisons support the value of incorporating contextual features and show that LLM-based semantic enhancement provides benefits beyond direct feature encoding. The results also demonstrate the advantages of adaptive multimodal fusion over concatenation and the consistent contribution of semantic residual calibration to prediction accuracy.

\subsection{Hyperparameter Study}

\subsubsection{Sensitivity of Residual Weights}

\begin{figure*}[tb]
    \centering
    \vspace{-3mm}
    \includegraphics[width=0.96\linewidth]{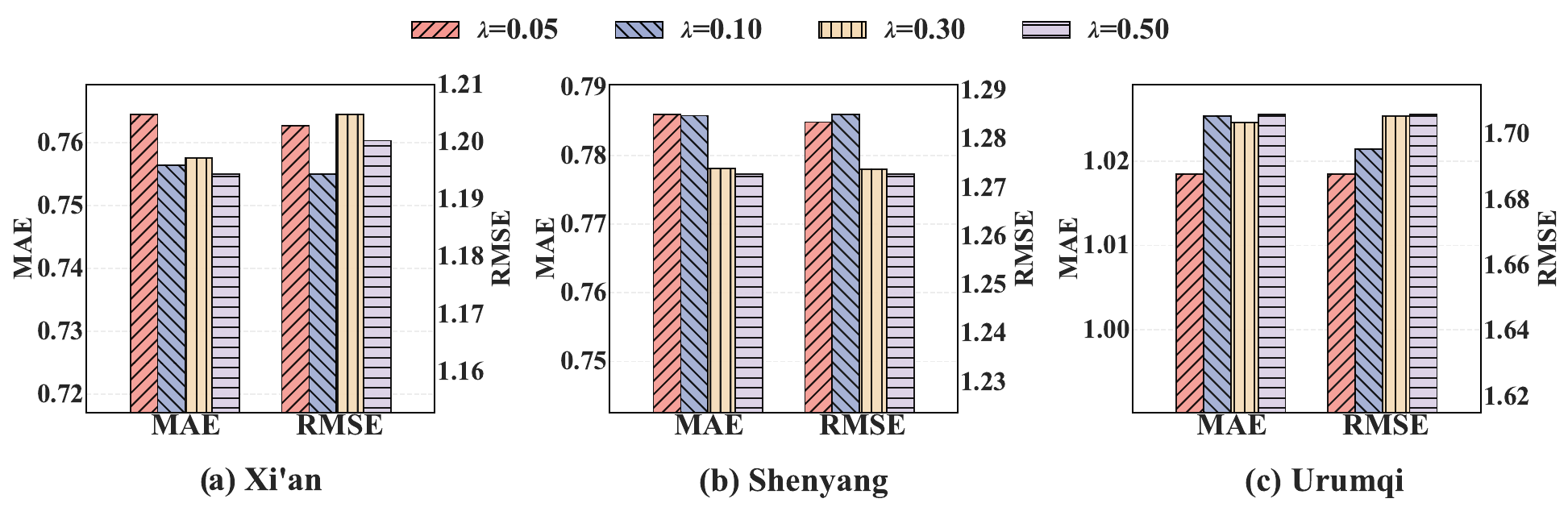}
    \vspace{-3mm}
    \caption{Sensitivity of the residual weight $\lambda$ in the Semantic-Enhanced Prediction Head across three cities.}
    \label{fig:residual_weight_sensitivity}
    \vspace{-3mm}
\end{figure*}

To assess sensitivity to the residual weight \(\lambda\) in the \textit{Semantic Enhanced Prediction Head}, we evaluate \(\lambda\in\{0.05,0.1,0.3,0.5\}\) while keeping all other settings fixed. Fig.~\ref{fig:residual_weight_sensitivity} presents the results.

Within each city, MAE varies by at most 1.3\% across the tested values. No single value performs best across all three cities. Xi'an and Shenyang achieve the lowest MAE at \(\lambda=0.5\), while Urumqi achieves the lowest MAE at \(\lambda=0.05\). These findings suggest that performance is relatively stable across the tested residual weights within \([0.05,0.5]\).

\subsubsection{Sensitivity to Historical and Prediction Lengths}

\begin{figure}[tb]
    \centering
    \vspace{-3mm}
    \includegraphics[width=0.96\linewidth]{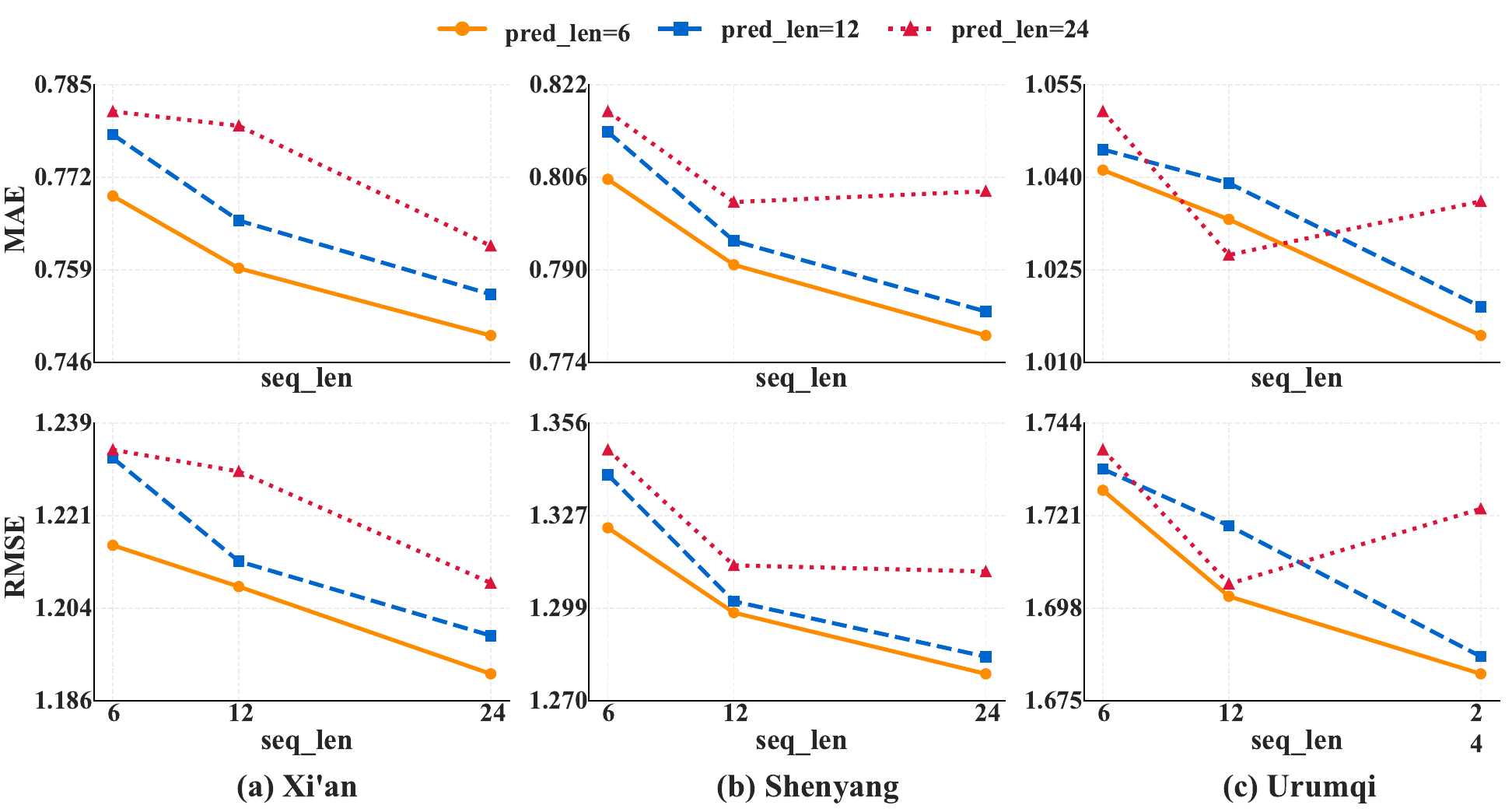}
    \vspace{-3mm}
    \caption{Sensitivity analysis over historical length (seq\_len) and prediction length (pred\_len).}
    \label{fig:sensitivity_seq_len}
    \vspace{-3mm}
\end{figure}

To assess sensitivity to historical input length (\texttt{seq\_len}) and prediction horizon (\texttt{pred\_len}), we vary each parameter independently over \(\{6,12,24\}\) hours, yielding a \(3\times3\) experimental grid. All other hyperparameters remain fixed. Fig.~\ref{fig:sensitivity_seq_len} summarizes the results.

Longer input sequences generally reduce MAE, while longer prediction horizons generally increase forecasting error. Across all three cities, the lowest MAE occurs with 24 hours of history and a 6-hour prediction horizon, yielding 0.7499, 0.7784, and 1.0143 for Xi'an, Shenyang, and Urumqi, respectively. The main experiments use the same historical window to predict the next 12 hours, yielding MAE values of 0.7556, 0.7826, and 1.0189.

The lower errors at a 6-hour horizon reflect an easier forecasting task. We therefore retain a 12-hour prediction horizon for the main comparisons and report the configuration with 24 hours of history and 6 hours of prediction as the best observed setting within the sensitivity grid.

\subsection{Sensitivity to Data Volume}

\begin{figure}[t]
    \centering
    \vspace{-3mm}
    \includegraphics[width=0.9\linewidth]{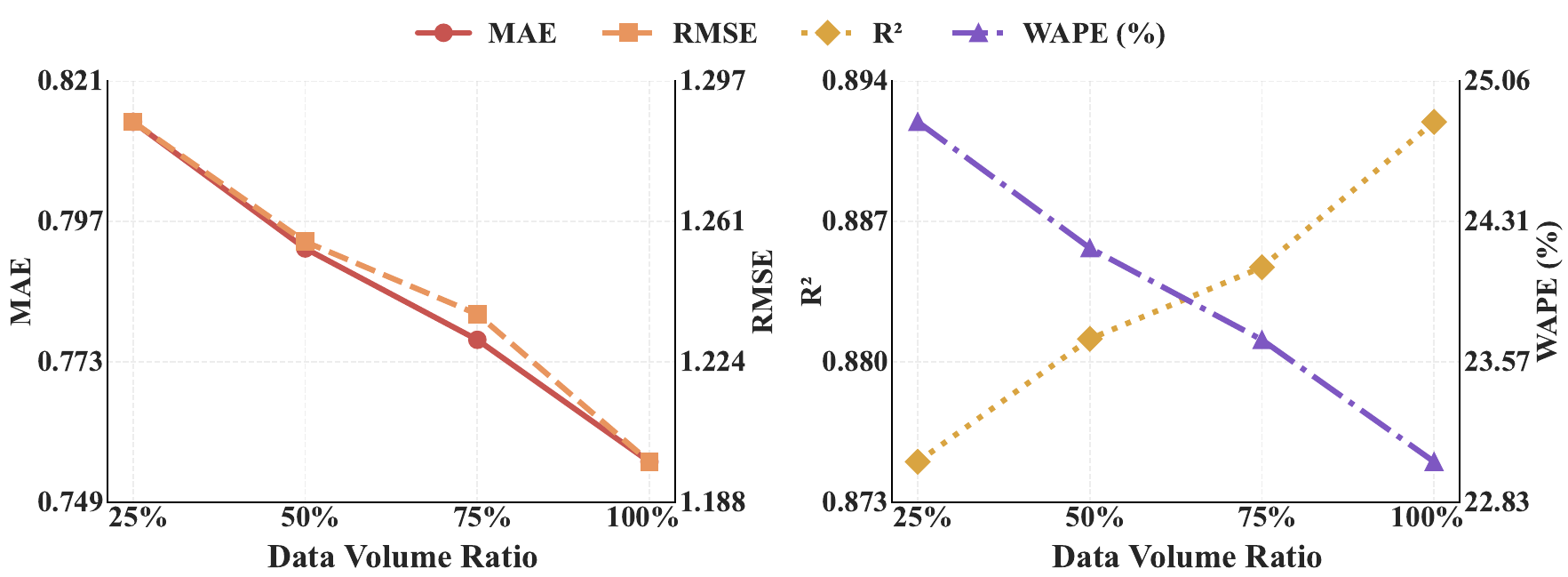}
    \vspace{-2mm}
    \caption{Sensitivity analysis of training data volume on model performance. Left: MAE and RMSE; right: $R^2$ and WAPE.}
    \label{fig:data_volume_sensitivity}
    \vspace{-2mm}
\end{figure}

To examine the effect of training data scale, we train Multi-STG on the Xi'an dataset using 25\%, 50\%, 75\%, and 100\% of the training windows while keeping all other settings fixed. As shown in Fig.~\ref{fig:data_volume_sensitivity}, performance improves as the training fraction increases. From 25\% to 100\%, MAE decreases from 0.8144 to 0.7556, RMSE from 1.2864 to 1.1982, and WAPE from 24.84\% to 23.05\%, while \(R^2\) increases from 0.8751 to 0.8917. These moderate but consistent gains suggest that Multi-STG benefits from additional training data, with gradual improvements across the evaluated data fractions.

\subsection{Cross-City Generalization}

\begin{figure*}[tb]
    \centering
    \vspace{-3mm}
    \subfigure[Multi-STG]{\includegraphics[width=0.48\linewidth]{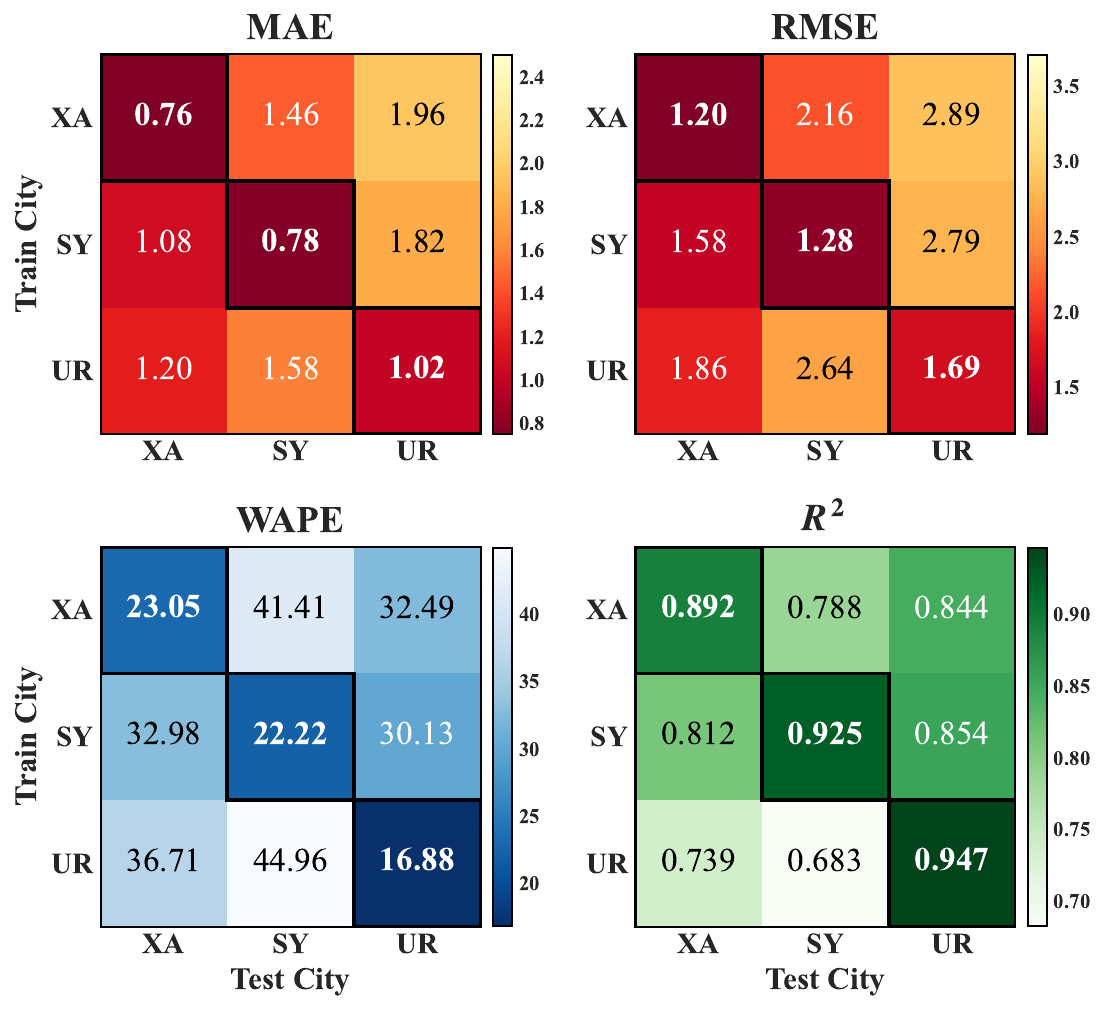}}
    \subfigure[DSTAGNN]{\includegraphics[width=0.48\linewidth]{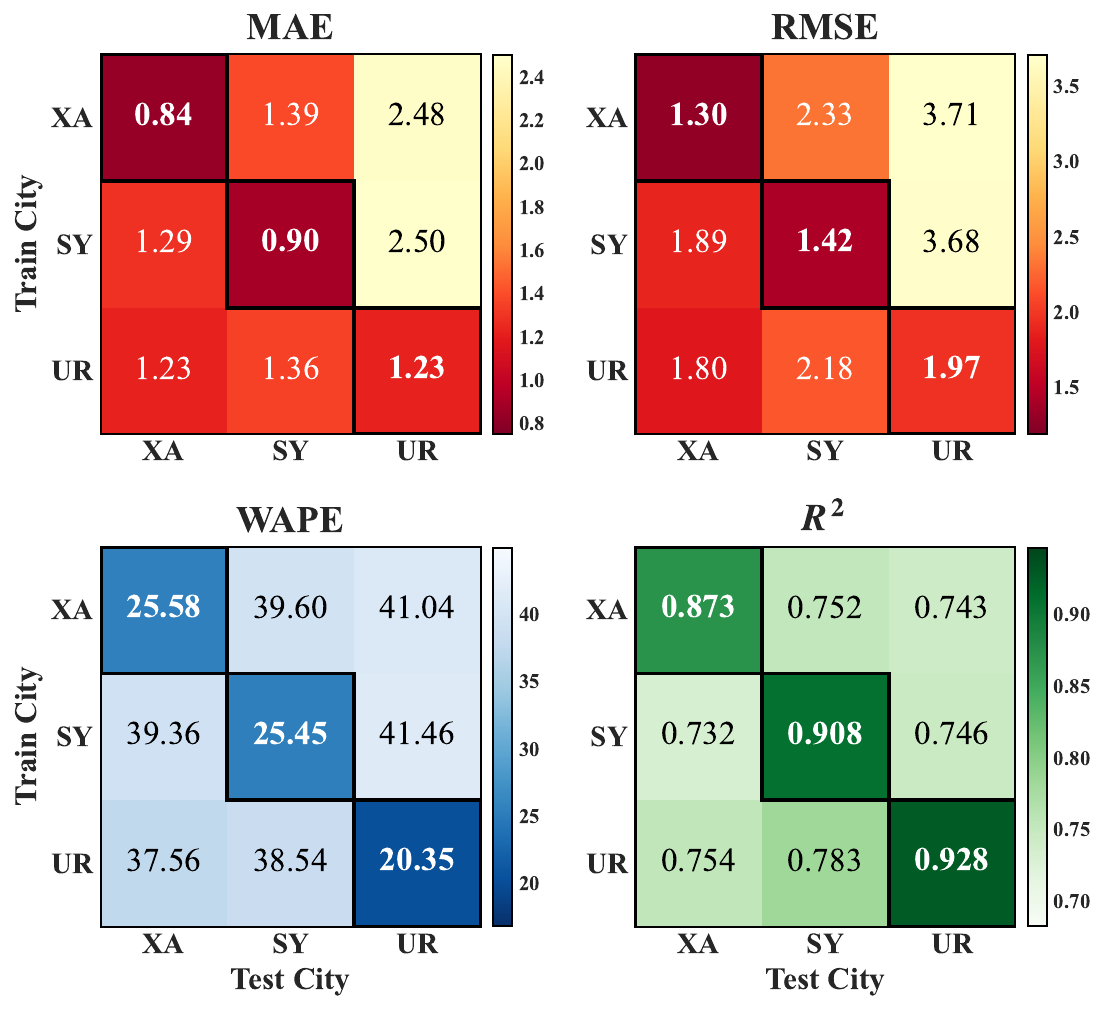}}
    \vspace{-3mm}
    \caption{Cross-city generalization. (a) Multi-STG. (b) DSTAGNN. Each subfigure reports MAE, RMSE, WAPE, and $R^2$. Within each heatmap, rows denote source cities and columns denote target cities; XA, SY, and UR denote Xi'an, Shenyang, and Urumqi, respectively. Colors are normalized per metric across both methods, and darker indicates better performance. The diagonal corresponds to in-city evaluation.}
    \vspace{-3mm}
    \label{fig:generalization_heatmap}
\end{figure*}

To assess transferability, we train Multi-STG on each source city and evaluate predictions on the other two target cities. Together with in-city evaluations, these experiments form a \(3\times3\) matrix. We compare Multi-STG with DSTAGNN, the strongest baseline on Xi'an. Normalization statistics are computed from the training segment of each target city, and both models follow the same data split and evaluation protocol. Fig.~\ref{fig:generalization_heatmap} reports MAE, RMSE, WAPE, and \(R^2\), using a shared color scale for each metric across both models. Diagonal entries reproduce the results from Table~\ref{tab:baseline}. For evaluations within Xi'an, Shenyang, and Urumqi, respectively, Multi-STG achieves MAE values of 0.76, 0.78, and 1.02 and RMSE values of 1.20, 1.28, and 1.69. The corresponding DSTAGNN results are 0.84, 0.90, and 1.23 for MAE and 1.30, 1.42, and 1.97 for RMSE. Multi-STG performs better across all four metrics in each city, with MAE reductions of 9.90\%, 12.68\%, and 17.03\%. For transfers between cities, MAE ranges from 1.08 to 1.96 for Multi-STG and from 1.23 to 2.50 for DSTAGNN. Across all nine source-target city combinations, Multi-STG performs better on seven combinations for each metric. DSTAGNN achieves lower MAE and WAPE when transferring from Xi'an to Shenyang, and better RMSE and \(R^2\) when transferring from Urumqi to Xi'an. For transfer from Urumqi to Shenyang, DSTAGNN performs better across all four metrics, achieving MAE of 1.36 versus 1.58, RMSE of 2.18 versus 2.64, WAPE of 38.54\% versus 44.96\%, and \(R^2\) of 0.783 versus 0.683. Averaged over all nine combinations, Multi-STG reduces mean MAE from 1.47 to 1.30 and mean RMSE from 2.25 to 2.01, indicating an overall advantage despite variation across transfer settings.